\documentclass[conference]{IEEEtran}
\pdfoutput=1
\IEEEoverridecommandlockouts
\usepackage{times}
\usepackage{pifont}
\usepackage[dvipsnames]{xcolor}
\usepackage{tikz}
\usepackage{makecell}
\usepackage{textcomp}
\usepackage[numbers]{natbib}
\usepackage{multicol}
\usepackage{graphicx} 
\usepackage{wrapfig}
\usepackage[export]{adjustbox}
\usepackage{subcaption}
\usepackage{threeparttable}
\usepackage{makecell}
\usepackage{multirow}
\usepackage{booktabs}
\usepackage{pifont}

\usepackage[ruled, lined, longend, linesnumbered]{algorithm2e}
\usepackage{amsmath}
\usepackage{amssymb}
\usepackage{amsfonts}
\usepackage{xcolor}
\definecolor{violet}{RGB}{148,0,211}
\usepackage[bookmarks=true]{hyperref}
\usepackage{url}

\begin{document}

\title{Few-Shot Neural Differentiable Simulator: Real-to-Sim Rigid-Contact Modeling}

\author{
    Zhenhao Huang$^{1}$,
    Siyuan Luo$^{2}$,
    Bingyang Zhou$^{1}$, 
    Ziqiu Zeng$^{1}$,
    Jason Pho$^{2}$ and
    Fan Shi$^{1,*}$
    \\$^{1}$ National University of Singapore, Singapore
    \\$^{2}$ Prana Lab, San Francisco, USA
    \thanks{$^*$ Corresponding author.}
}

\maketitle

\begin{abstract}
Accurate physics simulation is essential for robotic learning and control, yet analytical simulators often fail to capture complex contact dynamics, while learning-based simulators typically require large amounts of costly real-world data. 
To bridge this gap, we propose a few-shot real-to-sim approach that combines the physical consistency of analytical formulations with the representational capacity of graph neural network (GNN)-based models. 
Using only a small amount of real-world data, our method calibrates analytical simulators to generate large-scale synthetic datasets that capture diverse contact interactions. 
On this foundation, we introduce a mesh-based GNN that implicitly models rigid-body forward dynamics and derive surrogate gradients for collision detection, achieving full differentiability. 
Experimental results demonstrate that our approach enables learning-based simulators to outperform differentiable baselines in replicating real-world trajectories.
In addition, the differentiable design supports gradient-based optimization, which we validate through simulation-based policy learning in multi-object interaction scenarios. 
Extensive experiments show that our framework not only improves simulation fidelity with minimal supervision but also increases the efficiency of policy learning. 
Taken together, these findings suggest that differentiable simulation with few-shot real-world grounding provides a powerful direction for advancing future robotic manipulation and control. 

\end{abstract}
\section{Introduction}\label{sec:intro}

From basic grasping and placing to complex assembly and tool use, most manipulation tasks depend on intricate contact interactions that determine whether the operation succeeds or fails. Simulation provides a critical tool for allowing robots to learn and plan these tasks in a safe and controlled environment. Yet, despite decades of research in physics-based simulation, existing methods still struggle to accurately capture the contact dynamics observed in the real world.

\begin{figure}[t]
  \centering
   \includegraphics[width=\linewidth]{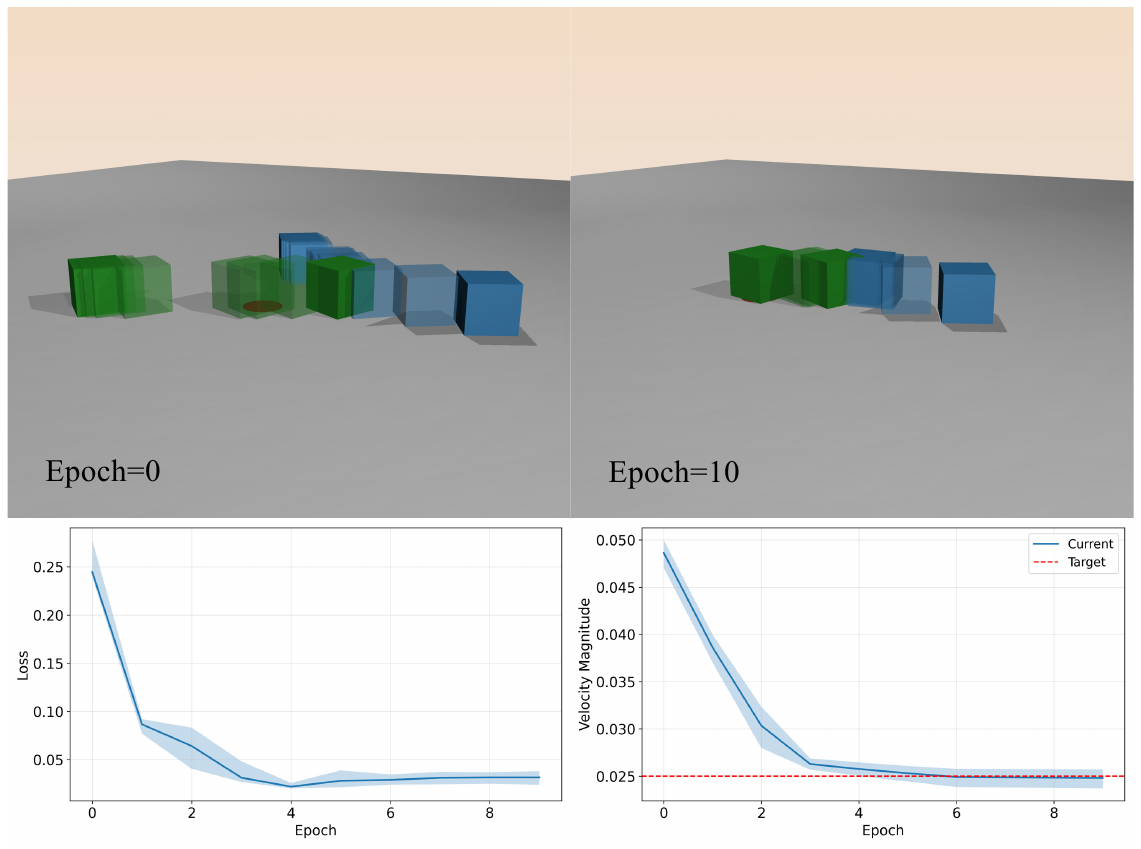}
   \caption{\textit{Top}: A blue cube is pushed to collide with a green cube, with the objective of stopping it at the red target area. The initial pushing velocity of the blue cube is optimized using stochastic gradient descent in our differentiable simulator, successfully achieving the goal. \textit{Bottom}: Convergence of optimization across multiple runs, showing the loss curve (left) and the magnitude of the initial pushing velocity (right).}
   \label{fig:gradient_based_optimization}
\end{figure}

In addition, these simulation approaches for robotic manipulation face a fundamental trade-off between physical accuracy and computational efficiency, particularly when modeling complex contact dynamics at scale. Traditional analytical simulators in robotics like IssacLab \cite{mittal2023orbit} and MuJoCo \cite{mujoco} focus on providing physically accurate and stable simulations but suffer from computational bottlenecks when scaling to complex scenes with numerous interacting objects. These simulators rely on constraint-based solvers that, while theoretically sound, become prohibitively expensive for contact-rich scenarios. Furthermore, such solvers are vulnerable and sensitive to parameters such as contact stiffness, damping, and coefficients of friction that are often difficult to measure in real-world scenarios.
Recently differentiable analytical simulators have emerged \cite{quentin2021differentiable,freeman2021brax,simon2023differentiable}, enabling gradient-based optimization for algorithms like motion planning and reinforcement learning \cite{song2024learning,luo2024residual}, but they still rely on the underlying contact modeling. 
Learning a portion of the simulation such as the signed distance and contact Jacobian \cite{pfrommer2020contactnets,bianchini2023simultaneous}, alleviate some of these issues. Recent advances in learnable simulators using GNNs have demonstrated promising results in achieving accuracy and computational efficiency, yet they require massive training data obtained via either cheap but inaccurate simulation or accurate but time-consuming real-world collection \cite{battaglia2016interaction,alvaro2018graph,li2018learning,pfaff2021learning,allen2023graph,allen2023learning}.

In this paper, we propose a novel framework that bridges the gap between real-world contact dynamics and learnable simulation. Our main contributions are as follows:
\begin{itemize}
  \item We propose a rigid-contact differentiable simulator based on GNN with surrogate gradients of nearest points with respect to object states for collision detection.
  \item We develop a few-shot real-to-sim data scaling pipeline with contact parameter identification to scale limited real-world data into large, diverse datasets with contact-rich interactions while maintaining physical realism.
  \item We show that our simulator outperforms the differentiable simulator Brax and achieves performance comparable to MuJoCo on real-world data, and highlight its potential for generalization to contact-rich scenes and gradient-based optimization in complex rigid-contact scenarios.
\end{itemize}

\section{Related Works}\label{sec:related works}

\subsection{Differentiable Physical Simulation}
Differentiable simulators have great potential in robotics, as they can be integrated with gradient-based optimization and learning algorithms for various tasks, such as system identification, trajectory optimization, and reinforcement learning \cite{song2024learning}. Recent works have proposed differentiable simulation for rigid-body dynamics \cite{howell2025dojodifferentiablephysicsengine,simon2023differentiable,freeman2021brax,quentin2021differentiable,eric2021neuralsim,simon2023singlelevel,murthy2021gradsim}. However, these methods often rely on analytical models that may not sufficiently capture real-world contact interactions. In contrast, our approach proposes a learnable GNN-based differentiable simulator that can adapt to real-world contact dynamics.

\subsection{Learnable Simulation}
Learnable simulators leverage data-driven methods to model the complex dynamics of physical systems, which are flexible in not only object types but also learnable parameters. The simulator can learn a residual physics neural network such as signed distance functions to model accurate rigid-contact simulation \cite{pfrommer2020contactnets,bianchini2023simultaneous,eric2021neuralsim}, or learn the entire dynamics using graph neural networks (GNNs) \cite{alvaro2018graph,allen2023graph,allen2023learning}. But large amounts of training data are required to train these models effectively. Our method addresses this by developing a few-shot data scaling pipeline which requires minimal real-world data.

\subsection{Real-to-Sim}
Some previous works have explored the real-to-sim approach to bridge the gap between simulation and reality. ContactNets \cite{pfrommer2020contactnets} utilize AprilTags \cite{krogius2019flexible} to obtain accurate 6D poses of objects from real-world images. GradSim \cite{murthy2021gradsim} employs differentiable simulation with differentiable rendering based on meshes to optimize physical parameters using simulated images, while DANO \cite{simon2023differentiable} uses neural density fields learned directly from real-world images for realistic simulation. We follow ContactNets, but with data scaling to reduce the amount of real-world data required for real-to-sim transfer.

\subsection{Collision Detection}
Collision detection is a crucial component of contact simulation. While the collision detection algorithms are not the focus of this work, the differentiability of collision detection is particularly relevant. There are several works on differentiable collision detection, either providing approximated gradients \cite{louis2023differentiable} or limited to convex primitives \cite{simon2022differentiable,kevin2023differentiable}. In our work, we derive the gradients of contact points or nearest points regardless of the differentiability of collision detection based on a reasonable assumption.

\begin{table}[t]
  \centering
  \caption{\label{tab:simulator_comparison}
    Comparison of our simulator against existing analytical and learned simulators.
  }
  \begin{tabular}{@{\hspace{6pt}}l@{\hspace{8pt}}c@{\hspace{8pt}}c@{\hspace{8pt}}c@{\hspace{8pt}}c@{\hspace{6pt}}}
    \toprule
    \textbf{Simulator} & \makecell{\textbf{Contact}\\\textbf{Model}} & \makecell{\textbf{Contact}\\\textbf{Accuracy}} & \textbf{Scalability} & \textbf{Differentiable} \\
    \midrule
    IsaacLab & PBD \cite{muller2007position} & Medium & High & \ding{55} \\
    MuJoCo & CCP & High & High & \ding{55} \\
    Brax & LCP & Low & Medium & \ding{51} \\
    ContactNets & DNN+LCP & Medium & Low & \ding{55} \\
    FIGNet & GNN & Medium & Medium & \ding{55} \\
    \textbf{Ours} & \textbf{GNN} & \textbf{High} & \textbf{High} & \ding{51} \\
    \bottomrule
  \end{tabular}
\end{table}
\section{Method}~\label{sec:method}
\begin{figure*}
  \centering
   \includegraphics[width=\linewidth]{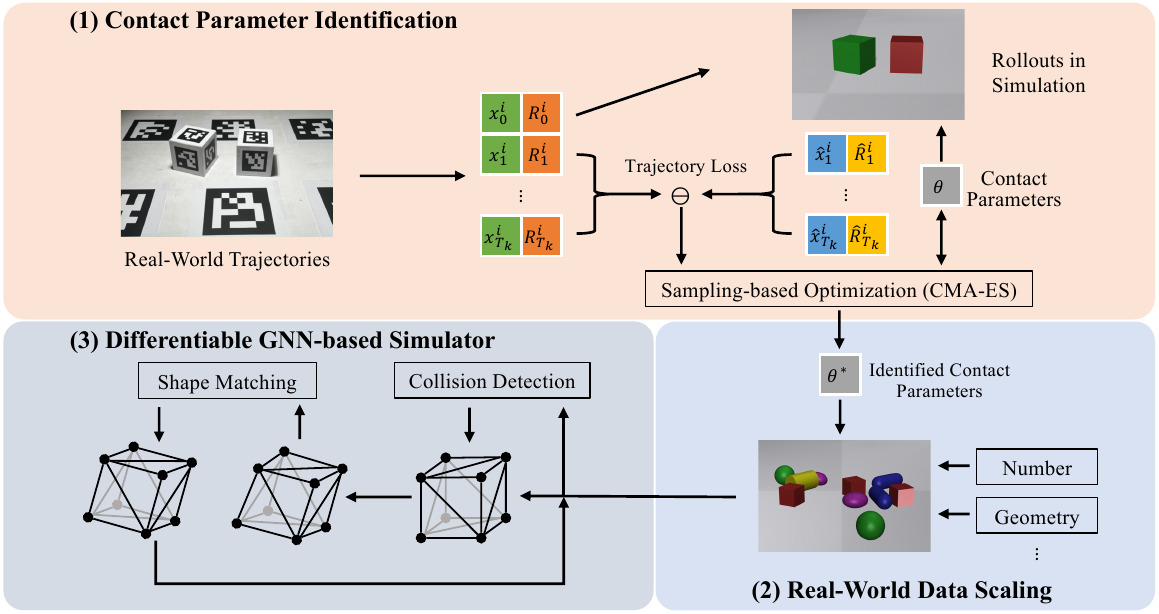}
   \caption{Illustration of our framework. (1) We utilize sampling-based identification of contact parameters to minimize the trajectory loss between real world and simulation, followed by (2) data scaling in terms of quantity, geometry, $etc.$ (3) The GNN is trained on the scaled dataset, equipped with differential collision detection and shape matching to achieve differentiability. The resulting simulator is capable of accurately modeling complex contact interactions.}
   \label{fig:framework}
\end{figure*}

The overall framework of our approach is illustrated in Figure \ref{fig:framework}. We begin by identifying key contact parameters that are critical for contact modeling but hard to measure directly (Section \ref{subsec:sysid}). Upon fixing identified contact parameters in the simulation, we proceed to scale the real-world data concerning the initial states and number of objects by generating a large dataset of diverse contact interactions (Section \ref{subsec:scaling}). Finally, we propose a fully differentiable GNN-based simulator that is well-trained by the scaled dataset (Section \ref{subsec:diffsim}).

\subsection{Contact Parameter Identification}~\label{subsec:sysid}
Given a set of collected real-world trajectories of dynamic rigid bodies, we formulate the contact parameter identification as an optimization problem. Let $\mathcal{D}=\{\mathcal{D}_k\}_{k=1}^K$ denote the collected dataset, where $\mathcal{D}_k=\{\{(\mathbf{x}_t^i,\mathbf{R}_t^i)\}_{i=1}^N\}_{t=0}^{T_k}$ is the set of positions and orientations of $N$ rigid bodies for the $k$-th demonstration that lasts for $T_k+1$ time steps. To obtain the missing velocity information, we use backward finite difference to compute the linear and angular velocities from the position and orientation trajectories. Similar to \cite{Brian2022validating}, the simulator starts from the same initial state as the real-world trajectory to get the corresponding simulated trajectory $\mathcal{\widehat{D}}_k(\boldsymbol{\theta})=\{\{(\mathbf{\widehat{x}}_t^i,\mathbf{\widehat{R}}_t^i)\}_{i=1}^N\}_{t=1}^{T_k}$.
The objective is to find the optimal contact parameters $\boldsymbol{\theta}^*$ that minimize the discrepancy between simulated and real-world trajectories:
\begin{equation}\label{eq:objective}
    \begin{split}
        \boldsymbol{\theta}^* = \arg\min_{\boldsymbol{\theta}} \sum_{k=1}^K \sum_{t=1}^{T_k} \sum_{i=1}^{N} (\frac{1}{w_i} \|\mathbf{x}_t^{i,k} &- {\mathbf{\widehat{x}}}_t^{i,k}\|_2 + \\
        |\text{Angle}&(\mathbf{R}_t^{i,k}, {\mathbf{\widehat{R}}}_t^{i,k})|)
    \end{split}
\end{equation}
where $w_i$ represents the positional scaling factor for the $i$-th body, and $\text{Angle}(\cdot)$ is the angular difference between the real and simulated orientations. 

In this work, we focus on identifying contact parameters in a simulator, using the high-fidelity physics simulator MuJoCo as our testbed. The contact solver in MuJoCo is primarily controlled by two parameter sets: \textit{solimp} and \textit{solref}. $\textit{solimp}=\{d_0,d_{\text{width}},\text{width},\text{midpoint},\text{power}\}$ defines the shape of the impedance function, mapping constraint violation to the generated force, while \textit{solref} specifies the time constant and damping ratio of the implicit spring-damper model. In addition, the lateral friction of coefficient $\mu$ is also a critical parameter for frictional contact interactions. These parameters $\boldsymbol{\theta}=\{\textit{solimp},\textit{solref},\mu\}$ govern the stiffness, dissipation, and stability of contact interactions, and are thus the primary targets of our contact parameter identification. Furthermore, since MuJoCo is non-differentiable, we employ a gradient-free optimization approach, CMA-ES \cite{akiba2019optuna}, to solve the optimization problem \eqref{eq:objective}. These identified parameters ensure that the subsequent data scaling produces physically realistic trajectories.

\subsection{Contact-Aware Data Scaling}~\label{subsec:scaling}
Training an accurate GNN-based simulator requires a large dataset of diverse contact scenarios \cite{pfaff2021learning,allen2023learning,allen2023graph}. However, constructing such a large real-world dataset is often expensive and time-consuming. Therefore, we propose to scale the scene by generating synthetic data using the contact parameters identified above. Unlike data augmentation techniques that apply transformations to real-world data to prevent overfitting, our data scaling method enriches the distribution of training data, thereby enhancing the GNN's ability to generalize to unseen scenarios. Since the identified MuJoCo is more physically realistic and stable for contact simulation, the distribution of scaled data is also relatively close to the real world, which can help our simulator better capture real-world contact dynamics. 

To scale data, we start with a base scene containing a few objects and systematically change the number and properties ($e.g.,$ geometry and mass) of objects with varying initial states in MuJoCo. This allows us to create a rich dataset that captures a wide range of contact interactions while maintaining realism.

\subsection{Differentiable GNN-based Simulator}~\label{subsec:diffsim}
Our GNN-based simulator is mainly composed of three parts: a collision detection algorithm, a GNN solver, and a shape matching module. The collision detection algorithm computes the minimum distance between objects for graph construction. The GNN solver predicts the accelerations of nodes and integrates them over time to update the positions, while the shape matching module ensures that the objects maintain their shapes.

\subsubsection{Mesh-based GNN solver}~\label{subsubsec:mesh_gnn}
Our GNN solver follows the FIGNet architecture \cite{allen2023learning}. The core idea is to represent each rigid body as a triangle mesh and model the dynamics through message passing on a graph where nodes correspond to mesh vertices and edges encode spatial relationships.

\textbf{Graph Construction.} For all objects in the scene, a graph is constructed containing two kinds of nodes (mesh and object nodes) and three kinds of bidirectional edges (mesh-mesh, object-mesh, and face-face edges).

\textbf{Message Passing.} The GNN utilizes the standard message-passing framework \cite{pfaff2021learning} with $L$ layers. At each layer $\ell$, nodes and edges are updated through:
\begin{align}
    \mathbf{e}_{ij}^{\ell+1} &= \phi_e^{\ell}(\mathbf{e}_{ij}^{\ell}, \mathbf{v}_i^{\ell}, \mathbf{v}_j^{\ell}) \\
    \mathbf{v}_i^{\ell+1} &= \phi_v^{\ell}(\mathbf{v}_i^{\ell}, \sum_{j} \mathbf{e}_{ij}^{\ell+1})
\end{align}
where $\phi_e^{\ell}$ and $\phi_v^{\ell}$ are learnable multi-layer perceptrons (MLPs), $\mathbf{e}_{ij}$ represents edge features, and $\mathbf{v}_i$ represents node features.

\textbf{Output.} The final decoder decodes the node features from the last message-passing layer and outputs accelerations for each mesh node:
\begin{equation}\label{eq:acceleration}
    \mathbf{a}_t^i = \phi_{\text{decoder}}(\mathbf{v}_i^{L})
\end{equation}
These accelerations are then integrated to update positions using a Verlet integrator $\mathbf{p}_{t+1}^{i} = \mathbf{a}_t^i + 2 \mathbf{p}_t^i - \mathbf{p}_{t-1}^i$. During inference, we implement the shape matching algorithm \cite{matthias2005meshless} to compute the rigid transformations for each object and project the predicted nodes to the transformed mesh instead of directly using the predicted node positions to keep the mesh topology consistent. Details of the GNN approach can be found in \cite{allen2023learning}.

\subsubsection{Surrogate Gradients of Collision Detection}~\label{subsubsec:diff_collision}
In our simulator, while the GNN solver is inherently differentiable, achieving end-to-end differentiability during inference requires making both collision detection and shape matching components differentiable. Shape matching involves standard matrix operations such as singular value decomposition and rotation matrix computations, which are readily supported by modern machine learning frameworks like \texttt{PyTorch} with automatic differentiation. Here we focus our attention on deriving gradients through collision detection, which presents a significant challenge as most of the collision detection algorithms either are non-differentiable or provide approximate gradients \cite{louis2023differentiable}.

We implement the discrete collision detection (DCD) via Coal \cite{coalweb}, a state-of-the-art collision detection library that provides efficient and robust collision detection algorithms over various geometric primitives. Since our GNN solver is mesh-based, the Gilbert-Johnson-Keerthi (GJK) algorithm \cite{GJK1988} and the Expanding Polytope Algorithm (EPA) \cite{EPA2001} are employed to compute the minimum distance or the penetration depth between convex meshes. However, these algorithms are non-differentiable due to their discontinuities. To address this, we derive the surrogate gradients that provides informative gradients and enables backpropagation through DCD.

Generally, given the states of two objects $ (O_A, O_B) $, the DCD algorithm outputs a set of contact pairs $ (o_A^i, o_B^j) $ together with the corresponding nearest points $ (p_A^i, p_B^j) $, where $ o_A^i \in O_A $ and $ o_B^j \in O_B $ denote the triangles of each object. By its very nature, DCD requires a distance threshold $ d_\epsilon $ to define potential contact pairs: if the distance between the nearest points of a triangle pair satisfies $ |p_A^i - p_B^j|^2 < d_\epsilon $, this pair is regarded as a potential contact. The choice of $ d_\epsilon $ is crucial. When the threshold is too small, certain true contacts may be missed. Conversely, when the threshold is too large, a substantial number of non-effective contacts are introduced, among which only a fraction will actually contribute in the subsequent contact response stage.

Consequently, by setting the threshold to a slightly larger range, we can make a reasonable assumption: due to the redundancy of detected contact pairs, the original set of contacts is sufficiently rich to support correct contact resolution, even when the motion undergoes small variations within the same time step. In this sense, the redundancy introduced by a larger threshold serves as a safeguard against local motion perturbations.

Returning to the differentiable simulation framework, and based on the above assumption, we regard the set of contact pairs as fixed within the current time step. Consequently, the requirement to differentiate through the contact detection process itself can be reasonably removed.

To obtain the gradients of the nearest points $ (p_A^i, p_B^j) $ with respect to the object states, we assemble the linear velocities of center of masses and angular velocities as $ u_A = [v_A^T, \omega_A^T]^T $ and $ u_B = [v_B^T, \omega_B^T]^T $, respectively. Similarly, the generalized positions are defined as $ q_A = [x_A^T, R_A^T]^T $ and $ q_B = [x_B^T, R_B^T]^T $, where $ x $ and $ R $ denote the position of the center of mass and the orientation expressed by quaternions of each object.
Then we define the kinematic mapping between the generalized velocities $ u = [u_A^T, u_B^T]^T $ and generalized positions $ q = [q_A^T, q_B^T]^T $ such that $ u = H \dot{q} $. Let $ r_A^i = p_A^i - x_A $ and $ r_B^j = p_B^j - x_B $ denote the relative positions from the center of masses to the nearest points, respectively.
Now we can get the relative velocities of the nearest points as:
\begin{equation}
    v_{\text{rel}}^{ij} = \begin{bmatrix}
      v_B + \omega_B \times r_B^j - (v_A + \omega_A \times r_A^i) \\
      v_A + \omega_A \times r_A^j - (v_B + \omega_B \times r_B^i)
    \end{bmatrix}
\end{equation}
We rewrite the above equation in matrix form as $ v_{\text{rel}}^{ij} = J^{ij} u = J^{ij} H \dot{q} $, where $ J^{ij} $ is the contact Jacobian matrix that maps the generalized velocities to the relative velocities of the nearest points. The contact Jacobian can be computed as:
\begin{equation}
    J^{ij} = \begin{bmatrix}
      -I & [r_A^i]_\times & I & -[r_B^j]_\times \\
      I & -[r_A^j]_\times & -I & [r_B^i]_\times
    \end{bmatrix}
\end{equation}
where $ I $ is the identity matrix and $ [\cdot]_\times $ denotes the skew-symmetric matrix. Next, we let $ p^{ij} = \begin{bmatrix} p_A^i \\ p_B^j\end{bmatrix} $. Finally, we can compute the surrogate gradients of the nearest points with respect to the generalized positions as:
\begin{equation}\label{eq:surrogate_gradient}
    \frac{\partial p^{ij}}{\partial q} = J^{ij} H
\end{equation}
The surrogate gradients in Eq. \eqref{eq:surrogate_gradient} can be efficiently computed once the contact pairs and nearest points are obtained from DCD.

Thus, our simulator is fully differentiable, supporting both forward simulation and backward gradient propagation.
\section{Experiments}~\label{sec:results}
\vspace{-6mm}
\subsection{Experimental Setup}
\begin{figure}[t]
  \centering
  \begin{subfigure}[b]{0.48\linewidth}
    \centering
    \includegraphics[width=\linewidth]{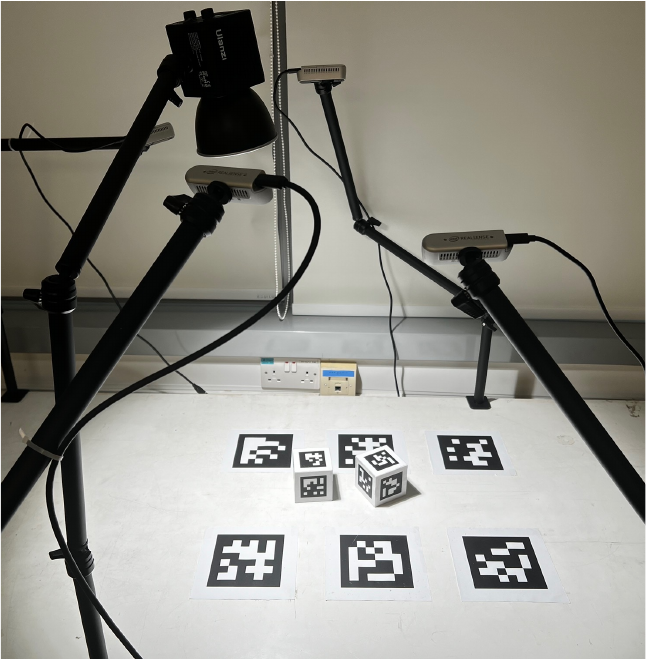}
    \caption{Real-World Data Collection}
    \label{fig:setup}
  \end{subfigure}
  \hfill
  \begin{subfigure}[b]{0.48\linewidth}
    \centering
    \includegraphics[width=\linewidth]{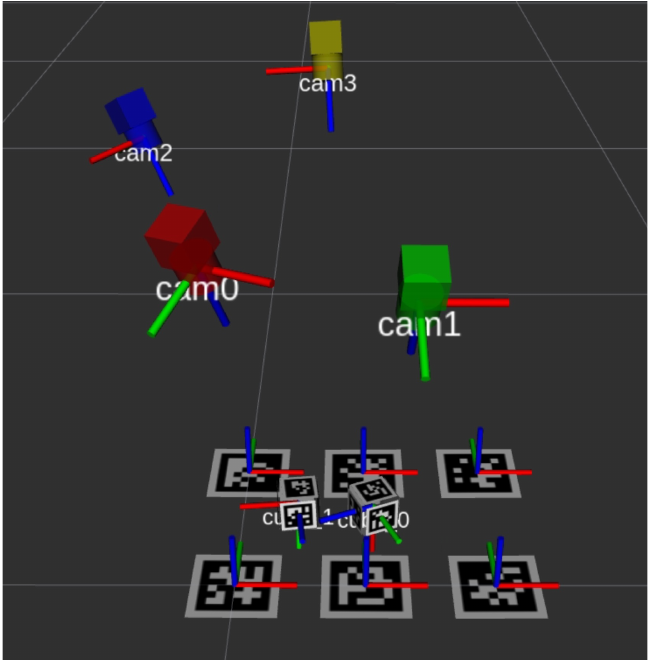}
    \caption{Pose Estimation}
    \label{fig:tagslam}
  \end{subfigure}
  \caption{(a) We collect real-world data on a tabletop setup with two cubes, considering a quasi-planar frictional contact scenario where one cube is pushed towards another cube at rest. (b) The poses of the cubes are estimated using TagSLAM.}
  \label{fig:data_collection_setup}
\end{figure}

Our real-world data collection system is shown in Figure \ref{fig:setup}. The cubes are 3D-printed and equipped with AprilTag \cite{krogius2019flexible} for robust fiducial detection. We utilize four Intel RealSense D435if cameras with $60$ Hz to detect the tags and estimate the cubes' poses via TagSLAM \cite{pfrommer2019tagslam}, shown as Figure \ref{fig:tagslam}. We collect only $3$ trajectories, each containing around $20$ frames, as the original training set. This small dataset is used for contact parameter identification of all analytical simulators in our comparison. We also augment it to $3000$ trajectories by rotating the trajectories around the z-axis for a learned baseline. Similarly, we use the identified MuJoCo to generate $3000$ trajectories as the training data for our GNN-based simulator. In addition, we collect another $14$ trajectories as the test set to evaluate the performance of all simulators. The initial positions and orientations of the cubes in the test set are different from those in the training set.
The whole framework is performed on a computer with AMD Ryzen 9 7950X3D CPU and single NVIDIA GeForce RTX 4090 GPU.

\subsection{Baselines}
We compare our GNN-based simulator with two analytical simulators, MuJoCo \cite{mujoco} and Brax \cite{freeman2021brax}. Brax is a popular differentiable simulator in robotics, which supports three built-in physics pipelines: (i) generalized pipeline resolves dynamics in generalized coordinates, similar to MuJoCo; (ii) positional pipeline uses position based dynamics to stabilize the simulation and resolve collision constraints fast; (iii) spring pipeline leverages an impulse-based method to enable a fast simulation. We also consider training our simulator directly on the real-world data with data augmentation (DA) as mentioned in Section \ref{subsec:scaling}, to evaluate the effectiveness of our data scaling.

\begin{table}[t]
  \renewcommand{\arraystretch}{1.25} 
  \centering
  \caption{Contact Parameters Optimization Range}
  \label{tab:contact_params}
  \begin{tabular}{ccc}
    \hline
    \textbf{\normalsize Simulator} & \textbf{\normalsize Parameter} & \textbf{\normalsize Range} \\
    \hline
    All Simulators & $\mu$ & $[0, 1]$ \\
    \hline
      & $d_0$ (\textit{solimp}) & $[0.9, 0.95]$ \\
      & $d_{\text{width}}$ (\textit{solimp}) & $[0.95, 0.99]$ \\
    MuJoCo  & Width (\textit{solimp}) & $[0.0001, 0.01]$ \\
    \& & Midpoint (\textit{solimp}) & $[0.001, 0.1]$ \\
    Brax (Generalized)  & Power (\textit{solimp}) & $[1, 5]$ \\
      & Time constant (\textit{solref}) & $[0.001, 0.1]$ \\
      & Damping ratio (\textit{solref}) & $[0.1, 10]$ \\
    \hline
    Brax (Positional \& Spring)  & Elasticity & $[0, 1]$ \\
    \hline
  \end{tabular}
\end{table}

\subsection{Contact Parameter Identification Results}
\begin{figure}[t]
  \centering
   \includegraphics[width=0.98\linewidth]{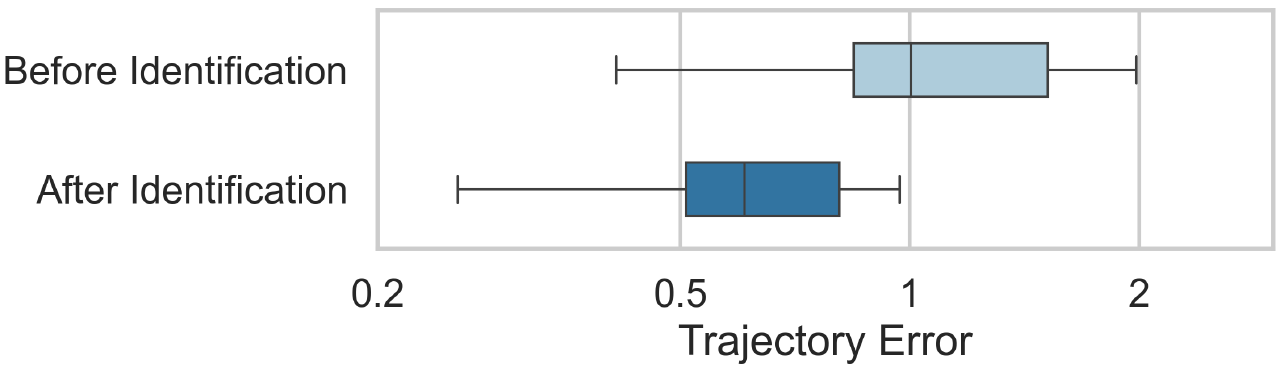}
   \caption{Comparison of trajectory errors (see Eq. \eqref{eq:objective}) in MuJoCo before and after contact parameter identification. The identified parameters significantly enhance the simulation's accuracy in replicating real-world contact dynamics.}
   \label{fig:sysid_box}
\end{figure}

\begin{figure}[t]
  \centering
   \includegraphics[width=0.98\linewidth]{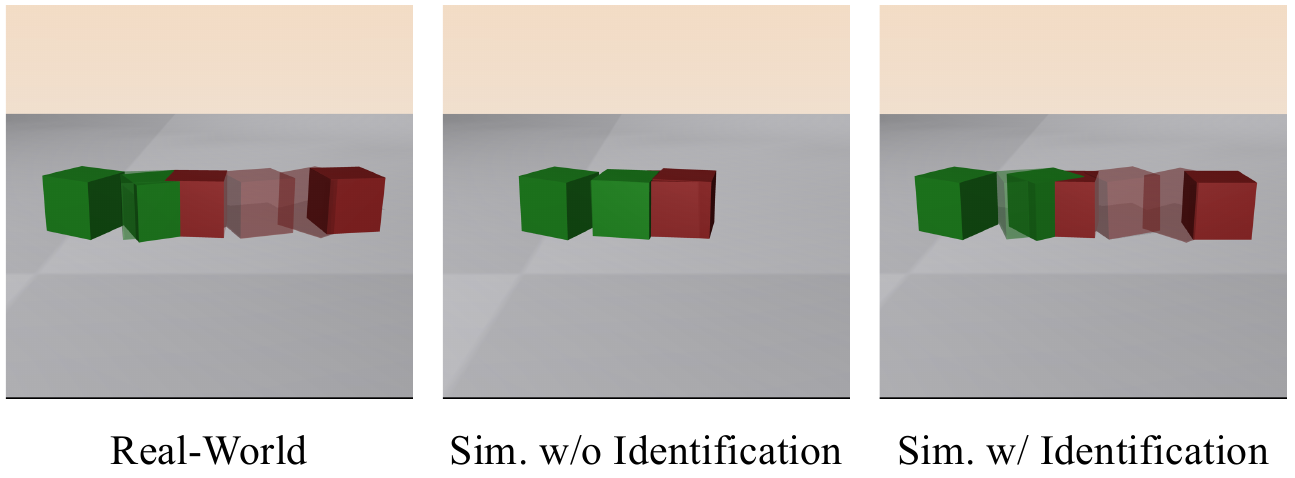}
   \caption{Illustration of a scenario where the moving cube (green) collides with the static cube (red) before and after contact parameter identification.}
   \label{fig:sysid_viz}
\end{figure}

We implement the sampling-based method CMA-ES to identify the contact parameters of MuJoCo through \texttt{Optuna} Python library \cite{akiba2019optuna}. The optimization range of each contact parameter is shown in Table \ref{tab:contact_params}. The comparison results of MuJoCo are shown in Figure \ref{fig:sysid_box}, where the average trajectory error on the test set is significantly reduced from $1.14$ to $0.73$ after contact parameter identification. Since MuJoCo has well-established physical models and formulates the contact dynamics in a convex unconstrained optimization problem, the few-shot identification is effective and efficient for MuJoCo to align with real-world contact dynamics. In Figure \ref{fig:sysid_viz}, we visualize the trajectories simulated in MuJoCo before and after contact parameter identification to the real-world trajectory, with the same initial states. Our identification process significantly improves MuJoCo's modeling of real-world contact dynamics, ensuring that the simulation environment more accurately reflects actual physical interactions.
However, there is still a gap between the identified MuJoCo and the real-world data, which indicates that the well-established analytical models in MuJoCo could not fully reflect the complexities of real-world contact interactions.

\subsection{Simulation Evaluation}

\begin{figure*}
  \centering
  \begin{tikzpicture}
    \node[anchor=west] at (0,0) {
      \begin{subfigure}[b]{0.27\linewidth}
        \centering
        \includegraphics[width=\linewidth]{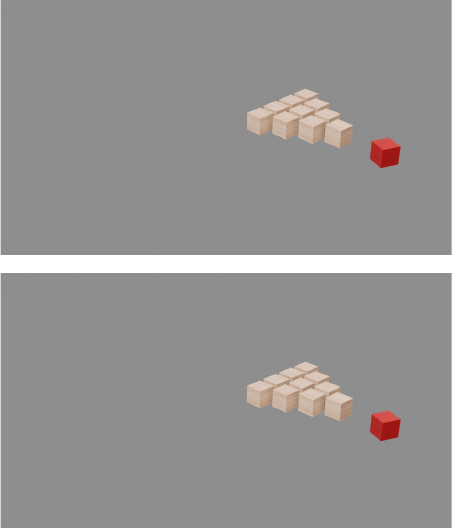}
        \caption{Frame $0$}
        \label{fig:rollout_0}
      \end{subfigure}
      \hfill
      \begin{subfigure}[b]{0.27\linewidth}
        \centering
        \includegraphics[width=\linewidth]{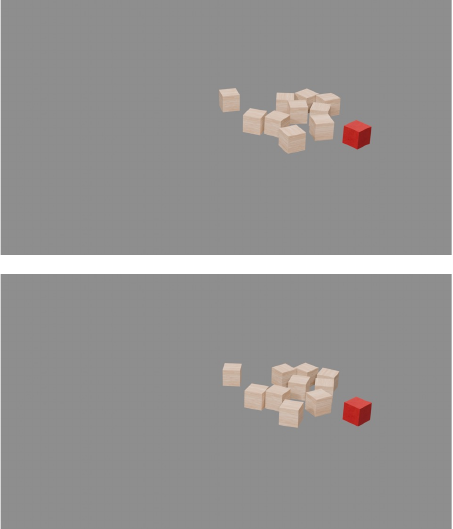}
        \caption{Frame $10$}
        \label{fig:rollout_10}
      \end{subfigure}
      \hfill
      \begin{subfigure}[b]{0.27\linewidth}
        \centering
        \includegraphics[width=\linewidth]{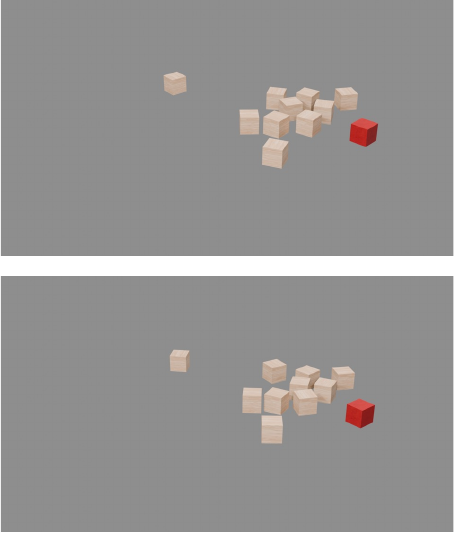}
        \caption{Frame $20$}
        \label{fig:rollout_20}
      \end{subfigure}
    };
    \node[anchor=east, font=\small] at (0,1.5) {Ground Truth};
    \node[anchor=east, font=\small] at (-0.5,-1.2) {\textbf{Ours}};
  \end{tikzpicture}
  \caption{A rollout example from our GNN-based simulator, illustrating a cube is pushed and collides with a bowling-like array of ten cubes. The ground truth is the scaled data generated by the identified MuJoCo. The output trajectory of our simulator is close to the ground truth, showing that our simulator accurately captures the dynamics and interactions between objects.}
  \label{fig:gnn_multi_object}
\end{figure*}

\begin{figure}[t]
  \centering
  \begin{subfigure}[b]{\linewidth}
    \centering
    \includegraphics[width=\linewidth]{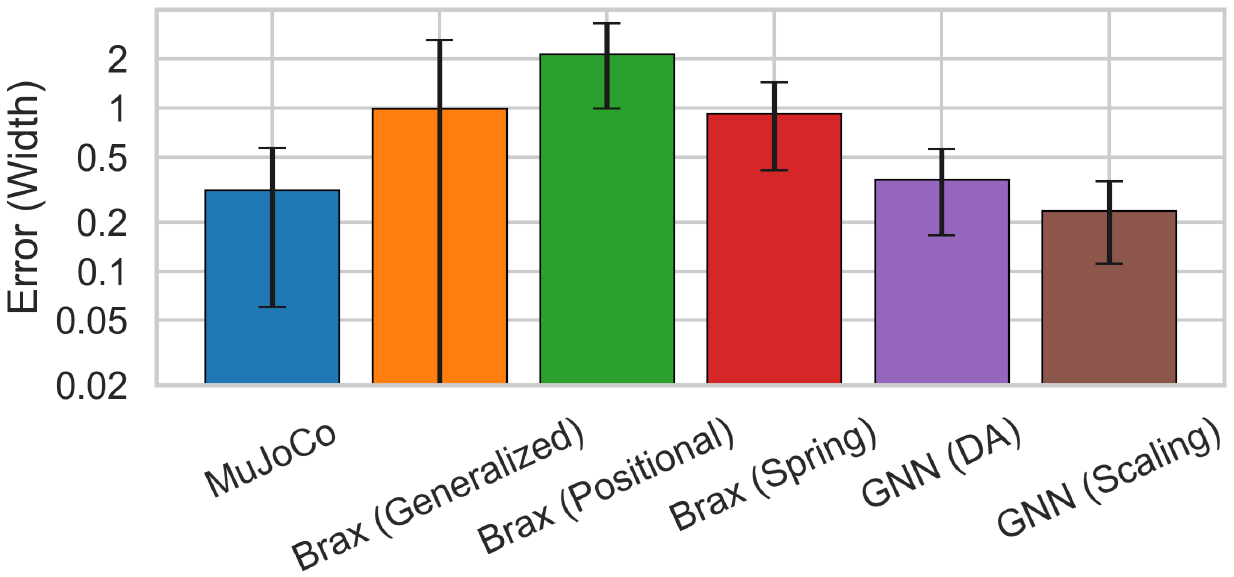}
    \caption{Positional errors}
    \label{fig:pos_error}
  \end{subfigure}
  \vfill
  \vspace{1mm}
  \begin{subfigure}[b]{\linewidth}
    \centering
    \includegraphics[width=\linewidth]{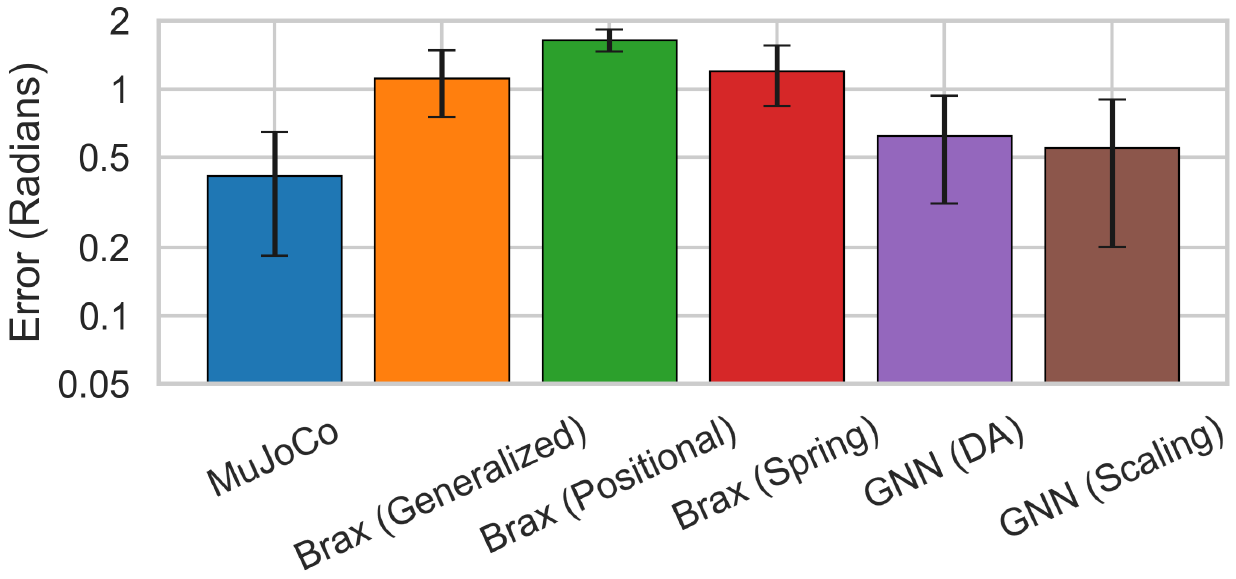}
    \caption{Angular errors}
    \label{fig:ang_error}
  \end{subfigure}
  \caption{The positional and angular errors of our GNN-based simulator against analytical and learned baselines. Our simulator achieves comparable performance to identified MuJoCo and outperforms all pipelines of identified Brax, while our data scaling method further enhances the accuracy of our simulator.}
  \label{fig:pos_ang_error}
\end{figure}

To validate the effectiveness of our GNN-based simulator, we evaluate the performance of our GNN-based simulator and all baselines on the test set. The positional and angular errors are shown in Figure \ref{fig:pos_ang_error}. Even though our simulator is trained only on the scaled data generated by the MuJoCo baseline, it still achieves similar error levels and even slightly lower positional errors compared to the identified MuJoCo. Thus, data scaling enables our simulator to learn the real-world rigid-contact dynamics well, thanks to the diverse configurations and relatively accurate contact modeling of the identified MuJoCo. Furthermore, both the positional and angular errors of our simulator trained with DA are higher than those of our simulator trained with the scaled data, which also demonstrates the effectiveness of our data scaling method. Compared to all pipelines of Brax, our simulator significantly outperforms them in both positional and angular errors. On the one hand, the identified Brax pipelines still have a large gap compared to the identified MuJoCo, indicating that Brax's physical models and contact formulations are less effective in aligning with real-world contact dynamics. On the other hand, our GNN-based simulator, with its data-driven nature and ability to learn complex interactions, can better model the rigid-contact dynamics from the scaled data.

We further conduct an experiment in which a single cube strikes a bowling-like array of ten cubes, shown in Figure \ref{fig:gnn_multi_object}, to evaluate the complex multi-body interaction capabilities of our simulator. The rollout results demonstrate that our simulator successfully captures near-instantaneous contact behaviors between cubes. This experiment highlights the ability of our GNN-based simulator to accurately simulate contact-rich and computationally heavy scenarios.

We showcase an example of gradient-based optimization in Figure \ref{fig:gradient_based_optimization}, where the initial pushing velocity of the blue cube is optimized to stop the green cube at a target area after colliding with it. The optimization successfully converges within $10$ epochs across multiple runs, demonstrating the potential of our differentiable simulator to enable gradient-based optimization for complex rigid-contact scenarios.

\section{Conclusion}~\label{Sec: Conclusion}
In this work, we propose a novel approach to learning a differentiable GNN-based rigid-body simulator from limited real-world observations. To enable effective training of our GNN simulator, we introduce a few-shot real-to-sim data scaling pipeline that identifies critical contact parameters of a physical simulator without the need of massive real-world data, and further generates a large synthetic dataset for our simulator to learn the diverse contact interactions from the calibrated simulator. Since collision detection is crucial for our simulator to model rigid-contact dynamics but non-differentiable, we derive the surrogate gradients of nearest contact points, allowing our GNN-based simulator to achieve full differentiability. Throughout the experiments, our data scaling alleviates the need for large amounts of real-world data, while it can still facilitate learning-based simulators to capture rigid-contact dynamics. Furthermore, our simulator shows the potential to generalize to novel multi-object interaction scenarios and support gradient-based optimization for contact-rich tasks.

The primary limitation of our approach is its heavy dependence on the accuracy of identification and contact modeling during real-world data scaling. Additionally, our method currently requires real-world object 6D poses, and assumes that the identified contact parameters generalize across different scenarios. In future work, more sophisticated contact representations that describe a broader range of contact dynamics can be investigated, along with the integration of our simulator with vision to enable direct learning from images or videos. Further evaluation with gradient-based control in closed-loop manipulation tasks would strengthen our simulator's practical utility in robotics.
\section*{Acknowledgments}
We acknowledge the support of MOE AcRF Tier 1 24-1234-P0001, Google Gift Fund, and NVIDIA Academic Grant Award.
We would also like to express our gratitude to Prana Lab for their valuable support and collaboration on this research.

\bibliographystyle{plainnat}
\bibliography{references}

\end{document}